\documentclass[10pt, a4paper]{article}

\usepackage[]{lrec-coling2024} 
\usepackage{microtype}
\usepackage{graphicx}
\usepackage{subfigure}
\usepackage{booktabs} 

\usepackage{amsmath}
\usepackage{amssymb}
\usepackage{hyperref}

\usepackage[T1]{fontenc}
\usepackage[scaled=0.85]{beramono}
\usepackage{listings}
\title{Biomedical Entity Linking for Dutch: Fine-tuning a Self-alignment BERT Model on an Automatically Generated Wikipedia Corpus}

\name{Fons Hartendorp$^1$, Tom Seinen$^2$, Erik van Mulligen$^2$, Suzan Verberne$^1$} 

\address{$^1$Leiden Institute of Advanced Computer Science, Leiden University, the Netherlands\\
$^2$Dept of Medical Informatics, Erasmus University Medical Center, Rotterdam, the Netherlands \\
\texttt{t.seinen@erasmusmc.nl}, \texttt{e.vanmulligen@erasmusmc.nl},       \texttt{s.verberne@liacs.leidenuniv.nl}\\
         \\}

\abstract{
Biomedical entity linking, a main component in automatic information extraction from health-related texts, plays a pivotal role in connecting textual entities (such as diseases, drugs and body parts mentioned by patients) to their corresponding concepts in a structured biomedical knowledge base. The task remains challenging despite recent developments in natural language processing. This paper presents the first evaluated biomedical entity linking model for the Dutch language. We use MedRoBERTa.nl as base model and perform second-phase pretraining through self-alignment on a Dutch biomedical ontology extracted from the UMLS and Dutch SNOMED. We derive a corpus from Wikipedia of ontology-linked Dutch biomedical entities in context and fine-tune our model on this dataset. We evaluate our model on the Dutch portion of the Mantra GSC-corpus and achieve $54.7\%$ classification accuracy and $69.8\%$ 1-distance accuracy. We then perform a case study on a collection of unlabeled, patient-support forum data and show that our model is hampered by the limited quality of the preceding entity recognition step. Manual evaluation of small sample indicates that of the correctly extracted entities, around 65\% is linked to the correct concept in the ontology. Our results indicate that biomedical entity linking in a language other than English remains challenging, but our Dutch model can be used to for high-level analysis of patient-generated text.
 \\ \newline  }

\begin{document}

\maketitleabstract

\section{Introduction}
\label{section:introduction}

\begin{figure}[t]
\vskip 0.2in
\begin{center}
\centerline{\includegraphics[width=0.4\textwidth]{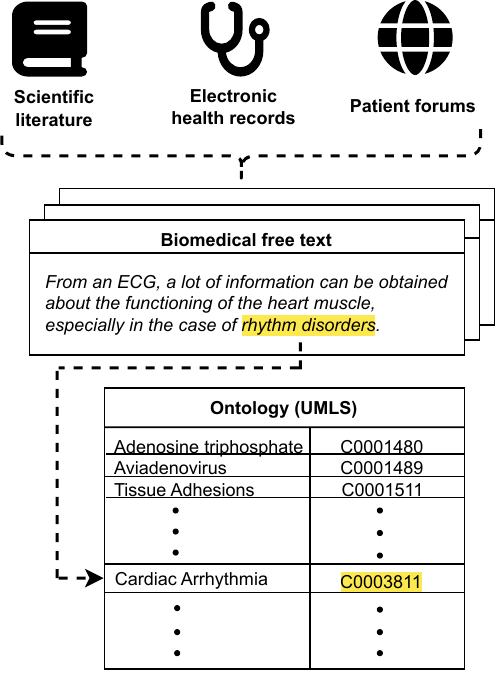}}
\caption{\small{The task of biomedical entity linking. An entity recognition model identifies entities in free text that are then passed to the biomedical entity linking (BEL) model. The BEL model associates the new, unseen mention with its corresponding concept from an ontology.}}
\label{fig:biomedical_entity_linking}
\end{center}
\vskip -0.2in
\end{figure}

Biomedical entity linking (BEL) is the task of linking mentions of biomedical entities in free text to their corresponding canonical form in a knowledge base \cite{garda2023belb} (Figure \ref{fig:biomedical_entity_linking}). 
Entity linking is a commonly used step after entity extraction to enable normalization and aggregation of entity mentions. Applications include automatically categorizing and improving search in medical scientific literature and information extraction from clinical notes and patient forums \cite{lee2016best}. In the analysis of patient experiences and patient--doctor communication, BEL can identify common concepts and aggregate free-text mentions from different authors and contexts. For example, a patient on an online support forum might mention that they have trouble with sleeping after taking medication. A BEL model would be able to link the mention ``trouble with sleeping'' to the medical concept \textit{insomnia} in a medical ontology and thereby aggregate all the mentions of insomnia from all patients.

Initial text pattern-based attempts to entity linking date back to the early 2000s, while modern models incorporate machine-learning algorithms \cite{french2022overview}. The task remains challenging for four reasons: 1) The high diversity in surface form of identical biomedical terms. For example, \textit{MI} and \textit{hartaanval} (heart attack) both belong to the same canonical concept form \textit{myocard infarct} (myocardial infarction). 2) The similarity in surface form of different biomedical terms: \textit{candida} and \textit{cardia} refer to a yeast and the heart respectively, while their Levenshtein distance is only two. 3) Free text generated by patients and medical professionals is often noisy, including spelling errors and (personal) abbreviations. 4) The number of entities in the biomedical domain is very large. The Unified Medical Language System (UMLS), the largest biomedical ontology and composed of various medical vocabularies, contains more than 3.3 million unique concepts \cite{bodenreider2004unified, vashishth2021improving}.

Labeled biomedical entity linking datasets are limited, particularly in languages other than English.\footnote{\url{https://paperswithcode.com/datasets?mod=texts&task=entity-linking}} In this paper, we present WALVIS, a weakly labeled Dutch biomedical entity linking dataset that was automatically generated using Wikidata and Wikipedia. 
We evaluate the quality of the WALVIS dataset and its effectiveness for training BEL models.
Specifically, we train a BEL model for Dutch using self-alignment pretraining of BERT (sapBERT) \cite{liu-etal-2021-self} on a cleaned Dutch sample of the UMLS. We further fine-tune this model in a supervised setting on WALVIS and evaluate it on the Dutch subset of the Mantra GSC corpus \cite{kors2015multilingual}. Additionally, we perform a case study on a collection of unlabeled, patient-support forum data to give an indication of the effectiveness of our Dutch SapBERT on patient-written texts.

The contributions of this paper are as follows: 1) introduction of a method for automatically generating a weakly labeled BEL-dataset in any preferred target language, by combining the UMLS, Wikidata and Wikipedia and thereby obviating the need for manual labelling by a domain expert. 2) introduction of the first evaluated BEL model trained on the Dutch language. 3) Evaluation of the model's performance and generalizability on the Dutch portion of the Mantra GSC dataset. 4) An analysis of the model's performance on patient-generated text through a case study on an online patient-support forum. We release our code and data on github.\footnote{\url{https://github.com/fonshartendorp/dutch_biomedical_entity_linking}}

\section{Related Work}
\label{sec:related_work}
The goal of BEL is to associate an entity mention in a text with its corresponding concept in a medical ontology, usually the UMLS. 
BEL models are commonly part of pipelines including biomedical named entity recognition (NER), followed by BEL and finally relation extraction \cite{french2022overview}. Some dedicated entity-linking corpora such as ShARe/CLEF and the NCBI dataset have been published \cite{pradhan2013task}, encouraging the development and evaluation of pure BEL-models without possible propagation of errors from the entity recognition module. A BEL-model typically involves a candidate generation step followed by candidate ranking \cite{mcinnes2009umls, d2015sieve}. 
In machine learning approaches to BEL, it is considered a \textit{mapping} problem. However, learning the mapping function is complicated by the lack of large, labeled datasets for training and the huge amount of classes \cite{loureiro2020medlinker}. 

With representation learning, the need for a labeled dataset can be obviated by leveraging the incorporated knowledge of a medical ontology. Since 2019, several BERT models \cite{devlin2019bert} for the biomedical domain have been released for English \cite{lee2020biobert, gu2021domain}. For BEL, the entity embeddings are then further improved in a second-phase pretraining step by using information from the ontology \cite{sung-etal-2020-biomedical, liu-etal-2021-self}. At inference, a similarity search is performed between the embedding of the new, unseen mention and the precomputed embeddings of all the terms from the ontology. The mention is then linked to the most similar term from the ontology. Self-alignment pretraining BERT (SapBERT) is a current state-of-the-art model that achieves $81.1\%$ accuracy on the COMETA corpus and $52.2\%$ on the MedMentions corpus \cite{liu-etal-2021-self, basaldella-etal-2020-cometa, loureiro2020medlinker}. Improvements have been attempted by incorporating context in the second-phase pretraining step or by using cluster-based inference \cite{zhang-etal-2022-knowledge, ujiie2021biomedical, angell-etal-2021-clustering}. In the past years, generative language models have also been explored for the task of BEL \cite{yuan2022generative}.

There is limited prior work on BEL for Dutch. There are two public medical annotation tools that includes Dutch and BEL: the rule-based Dutch implementation of MedSpaCy's QuickUMLS \cite{10.1093/jamia/ocad160}, and the Dutch model pack for the MedCat library \cite{kraljevic2021multi}. MedCat's linking module consists of two steps. First, dictionary matching is used for linking unambiguous terms (e.g. unique terms in the ontology, linked to one concept). Second, ambiguous terms are linked based on context embedding similarity. The context embeddings are trained on unambiguous terms and their context, in this case from the Dutch medical Wikipedia articles. Although MedCat uses similarity search with Word2Vec or BERT embeddings, the embeddings are not refined by leveraging knowledge incorporated in the ontology.

\section{Preliminaries}
\label{sec:background}

BEL is the task of mapping entity mentions in text documents to canonical concepts in a given ontology. A mention is a string that describes an entity in natural language. A concept is semantic unit that is clearly defined in the ontology and has a unique identifier. Mentions (words/phrases from text) can refer to either real world entities or abstract concepts from the ontology. We formally define the task of BEL as follows:

\textbf{Problem definition} Given a biomedical ontology $O$ consisting of $n$ concepts $O = \{c_1, c_2, ... , c_n\}$, a document D that contains a set $M$ of $p$ biomedical mentions $M=\{m_1, m_2, ..., m_p\}$, the task of BEL is to learn a mapping $M \rightarrow O$ that maps the mention $m_j \in M$ to the corresponding concept $c_i \in O$ that it refers to. 

\subsection{Unified Medical Language System}
The Unified Medical Language System (UMLS) is a large and comprehensive biomedical ontology created and maintained by the US National Library of Medicine. It is a collection of over 160 vocabularies, containing more than 15 million entries in 27 different languages. It maps entries from different databases and terminologies to around $3.3$ million unique concepts, that are identified by their Concept Unique Identifier (CUI). The Dutch portion contains around $290{,}000$ terms. The UMLS also contains data on $54$ types of semantic relations between concepts, both hierarchical (e.g. `is a') and non-hierarchical (e.g. `is conceptually related to').\footnote{\url{https://www.nlm.nih.gov/research/umls/knowledge\_sources/metathesaurus/release/statistics.html}} 

\subsection{Self-alignment pretraining BERT}
The main challenge of BEL in a representation learning setting is the quality of the entity embeddings \cite{basaldella-etal-2020-cometa}. Self-supervised learning with masked language modelling on medical data has improved BEL, but does not lead to a well separated representation space \cite{liu-etal-2021-self}. 

Self-alignment pretraining (sap)~\cite{liu-etal-2021-self} improves the embeddings of a pretrained BERT model, by self-aligning synonymous entries from a biomedical ontology. Formally, the goal of self-alignment is to learn a function $f(.;\theta) : O \rightarrow E$ that is parameterized by $\theta$ and where $O$ represents the set of terms in an ontology and $E$ the corresponding embedding representations with $\forall e \in E, \ e \in\mathbb{R}^d$. In sapBERT, $f$ is modelled by a BERT model with the output \texttt{[CLS]} token as embedding representation of the input term $c$. The similarity between two terms, $<f(c_i), f(c_j)>$ can be estimated by taking the cosine similarity. During the training procedure, \textit{online hard triplet mining} is used for generating informative pairs that are used for contrastive learning. From each mini-batch, a random anchor term $c_a$ is drawn. Together with a positive match -- or synonym -- $c_p$ and a negative match $c_n$, the triplet $(c_a, c_p, c_n)$ is formed. Informative triplets are generated by selecting positive matches (synonyms in the ontology) that get very dissimilar embeddings and, conversely, negative matches with embeddings that are nearly similar. Formally, triplets are selected that violate the following condition:
\begin{equation}\label{eq:triplet_condition}
        || f(c_a) - f(c_p) ||_{2} < || f(c_a) - f(c_n)||_{2} + \lambda 
\end{equation}
 where $\lambda$ is a pre-set margin. That is, we only select those triplets where the distance between the anchor and the positive term is larger than the distance between the anchor and the negative term plus margin $\lambda$. The mining of informative triplets only is useful for improving the embeddings, since otherwise non-informative triplets would dominate the training process due to the enormous size of the ontology \cite{liu-etal-2021-self}. The Multi-Similarity loss function is used for pulling the embeddings of positive pairs closer and pushing the embeddings of negative pairs further apart \cite{wang2019multi}. This process leads to a better separated representation space by leveraging the semantic biases of synonymy relations in the ontology.

\section{Methods}
\label{sec:methods}
Due to the need for expensive, manually labelling by domain experts, BEL datasets are not broadly available, especially in languages other than English. We introduce a method for automatically generating a weakly labeled BEL dataset in any given target language, by combining the structured knowledge source Wikidata, the UMLS and inter-article hyperlinks on Wikipedia. We implement the pipeline for Dutch. We first clean and enhance the Dutch subset of the UMLS and generate a Dutch biomedical ontology specifically tailored for BEL tasks. 

\subsection{Enhancing the UMLS}
\label{sec:enhancing_umls}

\begin{figure}[t]
\vskip 0.2in
\begin{center}
\centerline{\includegraphics[width=0.4\textwidth]{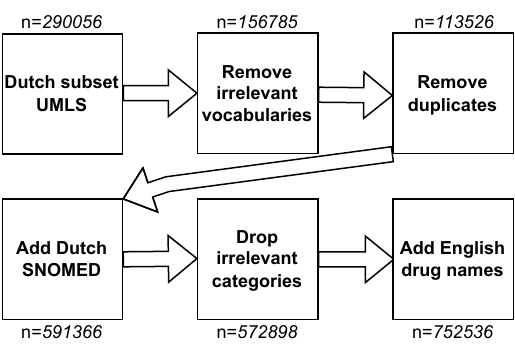}}
\caption{Flow diagram of ontology enhancement. The remaining number of entries are denoted in \textit{italic}.}
\label{fig:flow_diagram_ontology}
\end{center}
\vskip -0.2in
\end{figure}

Roughly $1.7\%$ of the UMLS 2022AB release, comprising $290{,}056$ terms, is in Dutch.\footnote{\url{https://www.nlm.nih.gov/research/umls/knowledge_sources/metathesaurus/release/index.html}} However, there is variability in the quality of the records. 
By following the same steps as the Dutch medical concepts project\footnote{\url{https://github.com/umcu/dutch-medical-concepts}}, we created a cleaned, UMLS-based Dutch biomedical ontology in several filtering and expansion steps.\footnote{Note that we cannot re-use their data because the UMLS is licensed and cannot be re-shared.} An overview is provided in Figure~\ref{fig:flow_diagram_ontology}. We exported, using the MetaMorphosys-tool, the terms of all seven Dutch vocabularies that are included in the release. 
From the $290{,}056$ terms, we removed the vocabularies \texttt{LNC-NL-NL} and \texttt{ICPC2ICD10DUT} that contain composed terms that are non-informative for the task of BEL such as \texttt{report:finding:date:polyclinical:document: endocrinology}.\footnote{Translated from Dutch for the readers' convenience} 
From the vocabularies \texttt{ICD10DUT} and \texttt{MDRDUT} we removed descriptive subterms, such as \texttt{non-specified}, as they are usually not found in free text. Also, duplicate entries were dropped, irrespective of capitalization. We added the Dutch SNOMED vocabulary, as this is not included in the UMLS. Since the US SNOMED is included in the UMLS, we matched Dutch to English terms on their SNOMED ID, and subsequently assign them their corresponding UMLS IDs (CUIs), 
dropping ambiguous terms. Entries linked to one of 26 semantic types that we considered non-relevant for BEL, such as \textit{Birds} and \textit{Geographic areas} were also removed. 
Finally, we added English drug names from the \texttt{ATC}, \texttt{DRUGBANK} and \texttt{RXNORM} vocabularies, since they are occasionally used in Dutch \cite{miller1995new, wishart2018drugbank, nelson2011normalized}.

The newly generated Dutch biomedical ontology contains $752{,}536$ terms sourced from 11 vocabularies, all linked to one of the $366{,}071$ distinct concepts. On average, each term is associated with one synonym, but the distribution is heavily right skewed (25\% percentile is 0 synonyms per term / 75\% percentile is 2 synonyms per term). Table \ref{tab:semantic_groups} shows the semantic group distribution of the ontology. The semantic groups are not classes in our entity linking problem, but rather a categorization of the classes. The four largest groups -- disorders (DISO), chemicals \& drugs (CHEM), procedures (PROC) and anatomy (ANAT) -- make up for $97\%$ of the terms in the ontology.


Since the UMLS and SNOMED are licensed, we cannot distribute the ontology. However, comprehensive details of all steps are provided in a Python Notebook in the project's Github repository. The ontology can be reproduced after requesting a UMLS and SNOMED license.


\begin{table*}[t]
\caption{Semantic group distributions of the ontology, train- and validation set of the no-duplicates, ontology-filtered subsets (*) from the WALVIS corpus (WALVIS*) and Mantra GSC (Mantra*). DISO: \textit{disorders}, CHEM: \textit{chemicals \& drugs}, PROC: \textit{procedures}, ANAT: \textit{anatomy}, LIVG: \textit{living beings}, PHEN: \textit{phenomena}, DEVI: \textit{devices}, PHYS: \textit{physiology} and ACTI: \textit{activities \& behaviors}, OBJC: \textit{Objects}, GENE: \textit{genes \& molecular sequences}, OCCU: \textit{occupations}, CONC: \textit{concepts \& ideas}. 1559 terms in the ontology are not assigned a semantic group (other). }
\label{tab:semantic_groups}
\vskip 0.15in
\begin{center}
\begin{small}
\begin{sc}
\begin{tabular}{llrrrrrrrrr}
\toprule
& & \multicolumn{2}{c}{\textbf{Ontology}} & \multicolumn{2}{c}{\textbf{WALVIS* tra.}} & \multicolumn{2}{c}{\textbf{WALVIS* val.}} & \multicolumn{2}{c}{\textbf{Mantra*}}\\
\textbf{Group} & \textbf{Example} & count & perc. & count & perc. & count & perc & count & perc. \\
\midrule
DISO & \texttt{MS} \textit{(multiple sclerosis)} & 310057 & 41.3 & 957 & 49.8 & 224 & 46.7 & 149 & 39.3\\
CHEM & \texttt{Neupro} & 185096 & 24.6 & 402 & 20.9 & 108 & 22.5 &  66 & 17.4 \\
PROC & \texttt{Dialyse} \textit{(dialysis)} & 124345 & 16.6 & 90 & 4.7 & 20 & 4.2 & 68 & 17.9\\
ANAT & \texttt{Heup} \textit{(hip)}& 108622  & 14.5 & 391 & 20.4 & 105 & 21.9 & 33 & 17.4 \\
LIVB & \texttt{Patiënt} \textit{(patient)} & 7586  & 1.0 & 14 & 0.7 & 6 & 1.2 & 29 & 7.7 \\
PHEN & \texttt{Licht} \textit{(light)}& 5997 & 0.8 & 4 & 0.2 & 1 & 0.2 & 7 & 1.8\\
DEVI & \texttt{IUD's} & 3153 & 0.4 & 3 & 0.2 & 0 & 0.0 & 5 & 1.3\\
PHYS & \texttt{Groei} \textit{(growth)}& 3125 & 0.4 & 33 & 1.7 & 11 & 2.3 & 19 & 5.0\\
ACTI & \texttt{Macht} \textit{(power)}& 1053 & 0.1 & 0 & 0.0 & 0 & 0.0 & 1 & 0.3 \\
OBJC & \texttt{Stof} \textit{(fabric)} & 678 & 0.1 & 13 & 0.7 & 3 & 0.6 & 2 & 0.5\\
GENE & \texttt{Codon} & 497 & 0.1 & 3 & 0.2 & 2 & 0.4 & 0 & 0.0\\
OCCU & \texttt{Genomics} & 464 & 0.1 & 10 & 0.5 & 0 & 0.0 & 0 & 0.0 \\
CONC & \texttt{Retentie} \textit{(retention)} & 304 & 0.0 & 0 & 0.0 & 0 & 0.0 & 0 & 0.0 \\
Oth. & & 1559 & 0.2 & 0 & 0.0 & 0 & 0.0 & 0 & 0.0 \\
\textbf{Total} & & \textbf{752536} & & \textbf{1920} & & \textbf{480} & & \textbf{379} & \\

\bottomrule
\end{tabular}
\end{sc}
\end{small}
\end{center}
\vskip -0.1in
\end{table*}

\begin{figure*}[ht]
\vskip 0.2in
\begin{center}
\centerline{\includegraphics[width=0.9\textwidth]{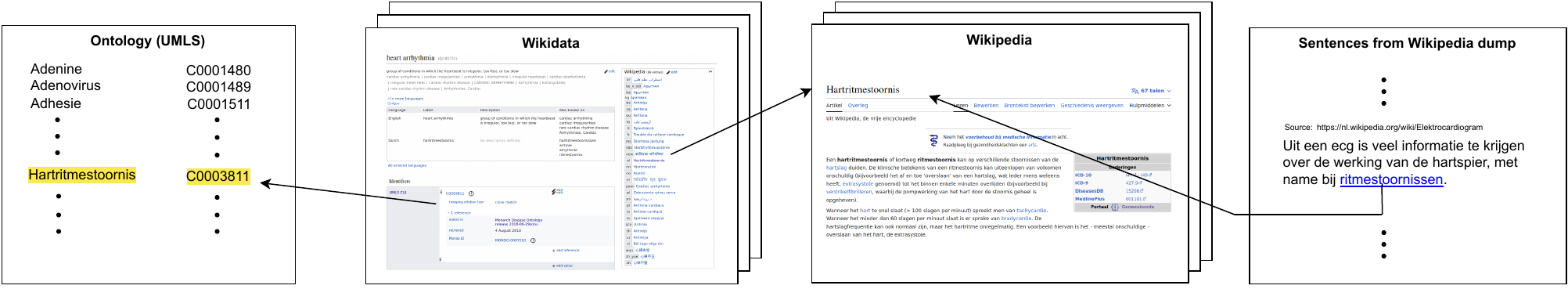}}
\caption{WALVIS corpus compilation. All Wikidata entries with a linked Dutch Wikipedia page and a UMLS CUI that are in the ontology are retrieved using SPARQL. Then, all sentences from the Wikipedia dump are parsed and selected if they contain a hyperlink to one of the collected Wikipedia pages.}
\label{fig:corpus_compilation}
\end{center}
\vskip -0.2in
\end{figure*}

\subsection{Corpus compilation}
\label{sec:corpus_compilation}
For the automatic generation of our weakly labeled dataset \texttt{WALVIS}, we combine our enhanced ontology with textual data from Wikipedia and structured data from Wikidata. An overview is provided in Figure~\ref{fig:corpus_compilation}. Wikidata is a collaboratively edited multilingual knowledge graph that acts as central storage for structured data of its Wikimedia sister projects including Wikipedia \cite{vrandevcic2014wikidata}. Relevant data can be obtained from Wikidata through SPARQL queries. We retrieve all $4,519$ Wikidata entities that have been assigned a UMLS CUI and are linked to a Wikipedia page that is part of the Dutch Wikipedia collection. The SPARQL query is provided in Listing~\ref{lst:sparql}.

\begin{lstlisting}[float=tp,captionpos=b, caption=SPARQL query for retrieving all Wikidata entities that contain a UMLS CUI\, and where there exists an article about the entity that is part of the Dutch Wikipedia, label=lst:sparql,
   basicstyle=\ttfamily,frame=single] 

SELECT ?concept ?conceptLabel ?cui ?article  WHERE {
  ?concept wdt:P2892 ?cui .
  ?article schema:about ?concept .
  ?article schema:isPartOf 
        <https://nl.wikipedia.org/>.

  SERVICE wikibase:label {
    bd:serviceParam wikibase:language "nl"
  }
}
\end{lstlisting}

We process all pages from the Dutch Wikipedia dump of March 2023\footnote{\url{https://dumps.wikimedia.org}} using the SpaCy sentence splitter with the Dutch \texttt{nl\_core\_news\_sm} pipeline. We then collect all $51{,}693$ sentences that contain a hyperlink to one of the $4{,}519$ Dutch Wikipedia articles that on their turn are linked to a Wikidata entity with a UMLS CUI property. The anchor texts of the hyperlinks are considered biomedical entity mentions. On average, a sentence contains $18$ $(\pm9)$ tokens and $53{,}960$ $(0.06\%)$ of the tokens in the collection are biomedical entity mentions that are linked to a UMLS CUI (Table \ref{tab:corpus_statistics}).  

For the WALVIS* subset, we kept each first unique mention and dropped their duplicates. Also, only mentions that link to a CUI that is present in the ontology are included. The WALVIS* corpus contains $2{,}400$ unique mentions from $2{,}307$ sentences. $1{,}751$ mentions are unseen by our model in the sapBERT training phase as they are not present in the ontology. The mentions map to $1{,}086$ unique CUI's that are all included in the ontology.

In Table \ref{tab:semantic_groups}, we see that the distribution of mentions over the semantic groups in the train- and validation set of WALVIS* is relatively similar to the distribution of terms in the ontology over the semantic groups, except for procedures (PROC). Procedures are possibly terms more commonly used by medical experts only, compared to disorders, chemicals \& drugs and anatomical terms, which could explain their lower prevalence on Wikipedia.

The code for parsing the Wikipedia dump and creating the corpus is available on Github, and the the WALVIS corpus and WALVIS* subset are available for download in XML format.

\begin{table}[t]
\caption{Corpora statistics. The WALVIS corpus contains many duplicate mentions that occur in different contexts. The WALVIS* subset (Wal.*) contains no duplicate mentions and only links to CUI's that have an entry in the ontology. We created a similar subset of the Mantra GSC corpus (Man.*).}
\label{tab:corpus_statistics}
\vskip 0.15in
\begin{center}
\begin{small}
\begin{sc}
\begin{tabular}{lrrr}
\toprule
 & \textbf{WALVIS} & \textbf{WAL.*} & \textbf{Man.*} \\
\midrule
Sentences & 51515 & 2307 & 166 \\
Avg. \#tok/sent & 18 & 20 & 17 \\
Mentions & 53781 & 2400 & 379 \\
Unique mentions & 3201 & 2400 & 379 \\
Unseen mentions & 49497 & 1751 & 214 \\
CUI's & 56141 & 2758 & 402\\
Unique CUI's & 1334 & 1086 & 359 \\
Unlinkable CUI's & 47548 & 0 & 0 \\
\bottomrule
\end{tabular}
\end{sc}
\end{small}
\end{center}
\vskip -0.1in
\end{table}



\subsection{Self-alignment pre-training} \label{sec:second_phase_pretraining}
We use the RoBERTa-derived language model MedRoBERTa.nl as base model. MedRoBERTa.nl was pretrained on nearly 10 million anonymized hospital notes obtained from the Amsterdam University Medical Centres \cite{verkijk2021medroberta}. The model is distributed with uninitialized head layers, allowing for fine-tuning on specific tasks.

We generate the training data for the self alignment pretraining from the cleaned Dutch medical ontology (Section~\ref{sec:enhancing_umls}). We generate a text file with positive pairs in the form of: \texttt{CUI||term 1||term 2}, where term 1 and term 2 are synonyms, so associated to the same CUI in the ontology. If more than 2 terms are associated to the same CUI, all pairwise combinations are traversed and added. We sample from the pool of positive pairs during the contrastive learning step for improving the pretrained BERT embeddings. Negative pairs are sampled online by randomly drawing a term from the ontology that is not linked to the same CUI. Both the negative and positive pairs must violate the minimum margin condition in Equation \ref{eq:triplet_condition}.

We use Multi-Similarity loss for re-aligning of the embeddings with parameters set to the same values as in \citet{liu-etal-2021-self}. 
We use a learning rate of 0.0001 with a weight decay of 0.01 for $\{0, 1, 3, 10\}$ epoch(s) with a batch size of 512. The similarity margin $\lambda$ is set to $0.2$. The \texttt{[CLS]} token is used as representation of the input term. The model is built in Pytorch 2.1.0, mostly based on code from \citet{liu-etal-2021-self}.\footnote{\url{https://github.com/cambridgeltl/sapbert}} 

\subsection{Fine-tuning}
Fine-tuning on the WALVIS corpus is performed in a similar manner. Now, the positive pairs are generated by combining mentions, linked terms and their corresponding CUI from the labeled dataset: \texttt{CUI||mention||linked term}. The hyperparameters are set to the same values as in Section \ref{sec:second_phase_pretraining}.
We fine-tune for $\{0,1,3,10\}$ epoch(s), building on the pretrained models from the previous step.




\subsection{Inference}
All terms from the ontology are fed to the trained model, generating a set of precomputed embeddings. At inference, a new, unseen mention is also fed to the trained model and a nearest neighbour search can be performed with the precomputed embeddings. The new mention is assigned the CUI of the most similar embedding from the ontology. Since a nearest neighbour search on $752{,}536$ items is computationally expensive, we built a FAISS index from the precomputed embeddings. FAISS is a library for approximate nearest neighbour search of dense vectors.\footnote{\url{https://github.com/facebookresearch/faiss/wiki}} For memory purposes, the precomputed embeddings are first compressed by using only their first $256$ principal components.

\subsection{Evaluation data and metrics}
We evaluated our method on the Dutch subset of the Mantra GSC corpus. The Mantra GSC corpus is a hand-labeled corpus annotated by domain-experts that was originally created for biomedical concept recognition in languages other than English \cite{kors2015multilingual}. The texts are sourced from MEDLINE titles and drug labels. The biomedical entities are also annotated with a UMLS CUI, that we use as gold labels for our linking model. Since the ontology does not contain all UMLS CUIs, we use the WALVIS* and Mantra* subsets that contain only mentions that link to a CUI that is included in the ontology. In both corpora, duplicate mentions were also removed since our model is not context-aware. Table \ref{tab:corpus_statistics} shows the corpora statistics of WALVIS* and Mantra*. The Mantra* subset contains 379 mentions from 166 sentences. The sentences are slightly shorter than the WALVIS* sentences, on average 17 tokens per sentence, and have more entity mentions per sentence. 


For finding the optimal number of sapBERT- and fine-tune epochs, we performed a hyperparameter optimization on the train set of WALVIS* and validated on its validation set. In the evaluation phase, we fine-tuned our optimal model on the full WALVIS* subset and evaluated on the Mantra* corpus. 

In addition to our primary metric classification accuracy, we also look at the 1-distance accuracy. For this metric, predictions are scored correct if they are any kind of 1-distance UMLS relation away from the gold label. For example, the prediction \textit{cystopyelonephritis} for a term with gold label \textit{pyelonephritis} would be correct since the UMLS contains a `classified as'-relation between the two.

\subsection{Case study on patient-support forum}
The Dutch online patient-support forum \url{https://www.kanker.nl/} donated anonymized textual data from between 2013 and 2016 in the form of blog posts, discussions and question-answering threads. The data does not have any manual annotations. We split the data in sentences using pySBD's sentence splitter, finding $123{,}338$ sentences and $2{,}191{,}424$ tokens. Before being able to apply BEL, we need entity extraction. To that end, we finetune MedRoBERTa.nl for NER on the machine translated MedMentions dataset in Dutch \cite{seinen2024annotation}.\footnote{\url{https://github.com/mi-erasmusmc/DutchClinicalCorpora}} This gives us $368{,}840$ medical named entities. We run both the  base model and our finetuned BEL model on these entities to link them to the Dutch UMLS and analyze the results.

\section{Results and analysis}
\label{sec:results}
We first assess the quality of our automatically generated WALVIS-corpus. We then turn to an evaluation of the optimal model on Mantra* and perform a brief error analysis. Finally, we explore its performance on entities sourced from the patient-support forum.

\subsection{Quality of WALVIS-corpus} We randomly sample 100 mentions from the WALVIS-corpus and manually evaluate the correctness of their label (Wikidata--UMLS link). The grading was performed by the first author using a tool that was developed for this purpose, for easy comparison of UMLS entries.\footnote{\url{https://anonymous.4open.science/r/biomedical_entity_linking-FCB4/ontology-browser/}} 
28 mentions were found to be linked to a concept that is related but not the same. For example, the mention \textit{kerndelingen} (divisions of the nucleus) on the Dutch Wikipedia page about asexual reproduction, is linked to \textit{cell nucleus}, which is related but not the same. The remaining 72 mentions seem to be labeled correctly. The label quality score indicates that the quality of the automatically generated corpus is suboptimal and that the data is not suited for evaluation purposes. The 100 samples and their grading can be found in our Github repository. 

\subsection{Main results} In hyperparameter tuning, the model with 3 sapBERT epochs and 10 fine-tune epochs performed optimal with a classification accuracy of $30.5\%$ and a 1-distance accuracy of $49.8\%$ on the validation set of WALVIS. Table \ref{tab:evaluation_results} shows the results on the Mantra* corpus. All results are averaged over 5 runs with different random seeds. The model (3S10FT) achieves a classification accuracy of $54.7\%$ and a 1-distance accuracy of $69.8\%$. That is a $10.1\%$ point and $13.1\%$ point improvement respectively, compared to the base model (BM).

In Table \ref{tab:evaluation_results}, the results grouped by semantic group are separately shown. We do not see a clear relation between the size of the semantic groups in the training data (the ontology and WALVIS), and their evaluation performance on Mantra*. On all four largest groups in the training data, an average improvement of around $10\%$ point in classification accuracy is observed, even though the largest group -- disorders (DISO) -- is with 149 samples four and a half times larger than the fourth-largest group ANAT (anatomy). We note that the numbers of mentions per semantic group in Mantra* are too small to derive clear conclusions.

\begin{table}[t]
\caption{Evaluation results on the Mantra* corpus for the base model (BM) and  our optimal model (trained for 3 self-alignment epochs + 10 fine-tune epochs). 
The semantic groups are not classes themselves, but rather a categorization of the classes. DISO: \textit{disorders}, PROC: \textit{procedures}, CHEM: \textit{chemicals \& drugs}, ANAT: \textit{anatomy}, LIVG: \textit{living beings}, PHYS: \textit{physiology}, PHEN: \textit{phenomena}, DEVI: \textit{devices}, OBJC: \textit{Objects}, ACTI: \textit{activities \& behaviors}. The total micro-average is shown for all 379 mentions averaged over 5 experiment runs with different random seeds.}
\label{tab:evaluation_results}
\vskip 0.15in
\begin{center}
\begin{small}
\begin{sc}
\begin{tabular}{lrrrrr}
\toprule
& & \multicolumn{2}{c}{\textbf{Accuracy}} &\multicolumn{2}{c}{\textbf{1-dist acc.}} \\
\cmidrule{3-6}
\textbf{Group} & \textbf{\#} & BM & 3S10FT & BM & 3S10FT \\
\midrule
DISO & 149 & 49.3 &  59.6 & 63.0 & 77.0 \\
PROC & 68 & 29.7 & 39.5 & 41.5 & 56.1 \\
CHEM & 66 & 48.2 & 57.6 & 58.5 & 67.3 \\
ANAT & 33 & 57.6 & 66.7 & 66.7 & 78.2 \\
LIVB & 29 & 33.8 & 48.3 & 48.3 & 61.4 \\
PHYS & 19 & 56.8 & 58.9 & 66.3 & 71.6 \\
PHEN & 7 & 57.1 & 76.2 & 71.2 & 82.4 \\
DEVI & 5 & 20.0 & 20.0 & 20.0 & 28.0 \\
OBJC & 2 & 0.0 & 0.0 & 50.0 & 70.0 \\
ACTI & 1 & 0.0 & 20.0 & 0.0 & 100.0\\
\midrule
\textbf{Total} & \textbf{379} & \textbf{44.6} & \textbf{54.7} & \textbf{56.7} & \textbf{69.8} \\ 
\bottomrule
\end{tabular}
\end{sc}
\end{small}
\end{center}
\vskip -0.1in
\end{table}

\subsection{Error analysis} We manually reviewed the mispredictions made by our optimal model. Due to the sometimes noisy and at some points extremely branched structure of the UMLS, seemingly small differences between prediction and gold label are scored incorrect. For example, the mention \textit{advies} (advice) is linked to \textit{voorlichting en advies} (counseling-C0010210) by our optimal model. However, in Mantra GSC, its gold label is given as \textit{adviseren} (advice-C0150600). The prediction is called correct by the 1-distance metric, since a RN (`Relation Narrow') exists between the two concepts in the UMLS. 

Sometimes, a mention is linked to an on surface form-level similar but semantically slightly different concept from the ontology. For example, mention \textit{cannabis} is linked to the plant genus \textit{cannabis} (C0936079), while its gold label in Mantra GSC is the drug \textit{cannabis} (C0678449). 
Also, the mention \textit{pijnlijke rug} (sore back) is linked to \textit{pijnlijke rug} (sore back-C0863105), but labeled as \textit{rugpijn} (back pain-C0004604). This indicates that our accuracies are perhaps an under-estimation of the actual effectiveness of the entity linking -- a finding that was also observed in previous work~\cite{dirkson2023others}.

We further observe a high focus on surface form by our model. For example, mention \textit{oren} (ears) is linked to \textit{ren} (running-C0022646) instead of gold label \textit{oor} (ear-C0013443). Moreover, mentions in all capitals, are often linked to a concept in all capitals,sometimes to a concept that is on surface form and semantical meaning very different. For example, mention \textit{SOMATOTYPE} is linked to \textit{DOPAMINERGIC AGENTS} (C0013036), while the surface form of its gold label is exactly similar to the mention but lower cased: \textit{somatotype} (C0037669). Lower casing all terms in the ontology and newly seen mentions could help, but by doing so some information is lost, for example in abbreviations (`pos' is commonly used for `positive', whereas `POS' could mean `Polycystic Ovary Syndrome'). While a context-aware model like KRISSBERT could reduce reliance on surface form, we did not implement such a model due to the lack of large, publicly available Dutch medical literature that is required for training \cite{zhang-etal-2022-knowledge}.

\begin{table}[t]
\caption{Top 5 most found named entities and corresponding linked concepts from the patient-support forum data from \url{kanker.nl}. Translated from Dutch for the reader's convenience.}
\label{tab:kankernl_results}
\vskip 0.15in
\begin{center}
\begin{small}
\begin{sc}
\begin{tabular}{lll}
\toprule
\textbf{\#} & \textbf{Named entity} & \textbf{Linked concept} \\
\midrule
4356 & cancer & primary malignant \\
 & & neoplasm \\
4240 & chemo & chemo-immunotherapy	\\
3043 & surgery & operative surgical  \\
 & & procedures \\
3034 & therapy & milieu therapy\\
2287 & tumor & neoplasms \\ 

\bottomrule
\end{tabular}
\end{sc}
\end{small}
\end{center}
\vskip -0.1in
\end{table}

\subsection{Case study} On the unlabeled data from \url{https://www.kanker.nl}, our finetuned BEL-model disagreed with the base model in the linking of $77.7\%$ ($286{,}654$) of the found named entities. This indicates that self-alignment pre-training and finetuning has a substantial effect on the model behaviour. To get an impression of the model quality, we manually graded 100 randomly sampled mentions. We found that 42 mentions were errors in the named entity recognition step. Of the 58 correct entity mentions, 20 entities (34\%) are linked to a wrong concept, another 20 mentions (34\%) are linked to a related concept and the remaining 18 (31\%) are linked correctly. The grading can be found on Github. 

The most commonly found named entities, such as \textit{kanker} (cancer), \textit{chemo} (chemo), and \textit{operatie} (surgery), seem likely to appear frequently on a support forum for cancer patients (Table \ref{tab:kankernl_results}). While the named entities are simpler terms compared to their linked concepts, for the majority they seem to be words that are also practiced by medical professionals. The main difference between layman talk and medical jargon is probably to be found in the context the words are used in. If we look at the semantic groups of the linked entities, we see that Disorders -- the largest group in the ontology -- is also the most used semantic group on the forum, but with $34\%$ much less proportionally than on Wikipedia ($50\%$ in WALVIS) and medical literature ($39\%$ in Mantra). Procedures on the other hand, are much more prevalent on the forum ($23\%$) than on Wikipedia ($5\%$) according to our model. 

\section{Conclusion}
\label{sec:conclusion}
To the best of our knowledge, our work is the first to introduce an evaluated biomedical entity linking model in the Dutch language. We also present a method for automatically generating a weakly labeled biomedical entity linking dataset in any preferred target language, by combining the data from a biomedical ontology, Wikidata and Wikipedia pages. Using this method, we introduce the first -- weakly labeled -- Dutch biomedical entity linking corpus: WALVIS. 
We trained a BEL model by self alignment pretraining on the MedRoBERTa.nl, followed by fine-tuning on WALVIS*. 
With around 70\% 1-distance accuracy on the external evaluation set Mantra*, 
we achieve a substantial improvement 
over the base model. This was achieved with a relatively small fine-tuning corpus. A case study on a collection of patient-written texts showed that the main source of error remains to be the named entity recognition step. Manual evaluation of small sample indicates that of the correctly extracted entities, our model links around 65\% to a correct or closely related concept in the ontology. 
In our evaluation on Mantra, we observe 
that our model relies heavily on surface form, which is for example observed by the erroneous linking of upper case mentions to upper case concepts that are otherwise very dissimilar. A context-aware model could further improve performance.

In conclusion, our biomedical entity linking model can be used for higher-level analysis of patient-oriented text data in Dutch. In future work, a larger corpus for fine-tuning could further improve the model's performance. A larger corpus could be created by automatically translating the English Wikipedia pages, which are not only larger in number, but also contain more words per article.

\newpage
\section*{Bibliographical References}\label{sec:reference}

\bibliographystyle{lrec-coling2024-natbib}
\bibliography{main}

\bibliographystylelanguageresource{lrec-coling2024-natbib}
\bibliographylanguageresource{languageresource}

\end{document}